\documentclass[letterpaper]{article} % DO NOT CHANGE THIS
\usepackage{aaai25}  % DO NOT CHANGE THIS
\usepackage{times}  % DO NOT CHANGE THIS
\usepackage{helvet}  % DO NOT CHANGE THIS
\usepackage{courier}  % DO NOT CHANGE THIS
\usepackage[hyphens]{url}  % DO NOT CHANGE THIS
\usepackage{graphicx} % DO NOT CHANGE THIS
\usepackage{natbib}  % DO NOT CHANGE THIS AND DO NOT ADD ANY OPTIONS TO IT
\usepackage{caption} % DO NOT CHANGE THIS AND DO NOT ADD ANY OPTIONS TO IT
\usepackage{algorithm}
\usepackage{algorithmic}
\usepackage{textcomp}
\usepackage{multirow}
\usepackage{pifont}
\usepackage{mathrsfs}
\usepackage{booktabs}
\usepackage{makecell}
\usepackage{mathrsfs}
\usepackage{diagbox}
\usepackage{bbding}
\usepackage{textcomp}
\usepackage{xcolor}
\usepackage{bm}
\usepackage{newfloat}
\usepackage{listings}
\usepackage{bbding}
\DeclareCaptionStyle{ruled}{labelfont=normalfont,labelsep=colon,strut=off} % DO NOT CHANGE THIS
\floatstyle{ruled}
\newfloat{listing}{tb}{lst}{}
\floatname{listing}{Listing}
\title{Category Prompt Mamba Network for Nuclei Segmentation and Classification}
\author{
    Ye Zhang\textsuperscript{\rm 1, \rm 2}, %\thanks{With help from the AAAI Publications Committee.},
    Zijie Fang\textsuperscript{\rm 3},
    Yifeng Wang\textsuperscript{\rm 4},
    Lingbo Zhang\textsuperscript{\rm 3},
    Xianchao Guan\textsuperscript{\rm 1, \rm 5},
    Yongbing Zhang\textsuperscript{\rm 1}\thanks{Corresponding author: Yongbing Zhang}
}
\affiliations{
    \textsuperscript{\rm 1} School of Computer Science and Technology, Harbin Institute of Technology (Shenzhen) \\
    \textsuperscript{\rm 2} Leibniz-Institut für Analytische Wissenschaften – ISAS – e.V.\\
    \textsuperscript{\rm 3} Tsinghua Shenzhen International Graduate School, Tsinghua University \\
    \textsuperscript{\rm 4} School of Science, Harbin Institute of Technology (Shenzhen) \\
    \textsuperscript{\rm 5} Pengcheng Laboratory \\
    
    zhangye94@stu.hit.edu.cn, ybzhang08@hit.edu.cn
}

\usepackage{bibentry}
\begin{document}

\maketitle

\begin{abstract}
Nuclei segmentation and classification provide an essential basis for tumor immune microenvironment analysis. The previous nuclei segmentation and classification models require splitting large images into smaller patches for training, leading to two significant issues. First, nuclei at the borders of adjacent patches often misalign during inference. Second, this patch-based approach significantly increases the model's training and inference time. Recently, Mamba has garnered attention for its ability to model large-scale images with linear time complexity and low memory consumption. It offers a promising solution for training nuclei segmentation and classification models on full-sized images. However, the Mamba orientation-based scanning method lacks account for category-specific features, resulting in sub-optimal performance in scenarios with imbalanced class distributions. To address these challenges, this paper introduces a novel scanning strategy based on category probability sorting, which independently ranks and scans features for each category according to confidence from high to low. This approach enhances the feature representation of uncertain samples and mitigates the issues caused by imbalanced distributions. Extensive experiments conducted on four public datasets demonstrate that our method outperforms state-of-the-art approaches, delivering superior performance in nuclei segmentation and classification tasks.
\end{abstract}


\section{Introduction}

With the advancements of whole-slide pathology image production and scanning technologies, pathological image analysis tasks represented by nuclei segmentation and classification \cite{ilyas2022tsfd, oh2023diffmix, zhang2024dawn} play a more and more critical role in cancer diagnosis \cite{kowal2014nuclei} and patient prognosis analysis \cite{wang2022cell}. The morphology and category of nuclei can reflect the cellular differentiation degree and distribution status, which provide valuable information for the tumor micro-environment analysis \cite{zamanitajeddin2024social}.

\begin{figure}[t]
	\centering
	\includegraphics[width=3.3in]{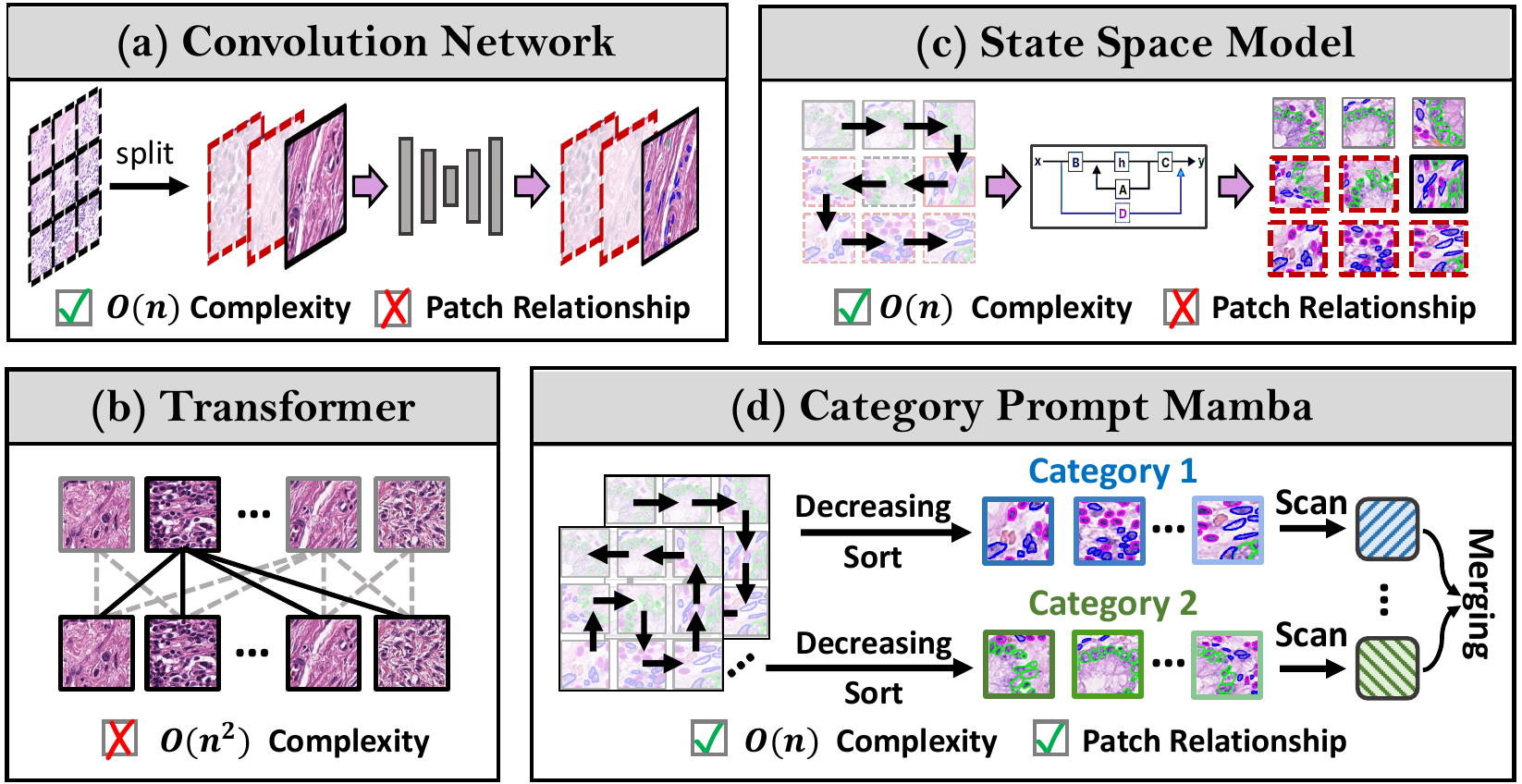}
	\caption{The existing nuclei analysis framework. The images with \textbf{black box} represent current training samples, and the images with \textcolor{red}{\textbf{red box}} do not participate in training. (a) represents convolution network independently trains each image; (b) represents Transformer structure has quadratic computational complexity; (c) represents state space model only considers the preceding samples; (d) is our proposed category prompt network, which utilizes the classification probability as a basis to guide the sequences sorting.}
	\label{fig: motivation}
        \vspace{-3mm}
\end{figure}

Recent developments in deep learning have enabled automatic nuclei segmentation and classification \cite{doan2022sonnet, pan2023smile}. However, due to the large size of pathological images, most existing nuclei analysis approaches require splitting these images into smaller, independent patches for training \cite{ronneberger2015u, chen2016dcan}, as illustrated in Fig.\ref{fig: motivation} (a). This patch-based training method presents two significant challenges. \textbf{First}, images often suffer from edge effect problems during inference, where nuclei at the borders of adjacent patches are misaligned. \textbf{Second}, when the distribution of nuclear categories is highly imbalanced within a single patch, the performance of rare categories tends to deteriorate \cite{sirinukunwattana2016locality, naylor2018segmentation}. This decline is primarily due to the insufficient training of rare categories, stemming from their limited sample sizes.

To address the patch misalignment issue in large-scale images, HoverNet \cite{graham2019hover} introduces a center feature clipping strategy to reduce the uncertainty associated with edge features. Similarly, DoNuSeg \cite{wang2024dynamic} employs an overlapped sampling strategy to mitigate edge effects. However, these approaches can merely alleviate misalignment; they do not fundamentally resolve the issue and often lead to increased model training and inference times. In tackling the class imbalance problem, some transformer-based methods are proposed \cite{he2023transnuseg, lou2024cell}. These approaches enhance the representation of rare categories through global feature interaction. By integrating information from distant regions, transformers can better contextualize rare categories, improving their prediction performance. However, the quadratic computational complexity inherent to the transformer architecture restricts the size of the input image and imposes substantial demands on memory resources, as illustrated in Fig. \ref{fig: motivation} (b). Additionally, although class-weighted loss functions are proposed to address class imbalance \cite{schmitz2021multi, hancer2023imbalance}, their impact on improving performance remains limited. These limitations underscore the need for more efficient and effective solutions to patch misalignment and class imbalance in nuclei segmentation and classification.

The state space model (SSM) represented by Mamba \cite{gu2023mamba} has recently garnered widespread attention for its ability to perform long-sequence modeling with linear complexity. Mamba's low memory consumption further enables training nuclei segmentation and classification networks directly on large-scale images, thereby eliminating the need for image splitting. However, Mamba's unidirectional scanning inherently limits interaction between patches, as it only considers preceding sequences and lacks awareness of subsequent ones, as illustrated in Fig. \ref{fig: motivation} (c).
To overcome the issue, several multi-directional scanning strategies are proposed. Vim \cite{zhu2024vision} implements forward and backward scanning to enhance interaction across all patches. In contrast, VMamba \cite{liu2024vmamba} introduces a four-directional cross-scan method that considers both horizontal and vertical directions. Additionally, other scanning strategies \cite{li2025mamba, yang2024plainmamba} are developed to more thoroughly address the influence of sequential order.
Despite these advancements, the current scanning strategies are orientation-aware rather than class-aware. In class imbalance scenarios, these methods often fail to adequately recognize and enhance class-specific features, leading to suboptimal performance. Hence, more sophisticated approaches are needed to address class imbalance in nuclei segmentation and classification effectively.

This paper addresses the challenges of class imbalance and large-scale image prediction by introducing a probability-guided sorting method built on the Mamba network. Our approach involves learning the feature representation of each category independently and then aggregating them, as depicted in Fig. \ref{fig: motivation} (d). This method enhances the feature representation of rare categories, thereby improving the accuracy of their predictions.
Specifically, we utilize category prompts as supervision to predict multi-category labels, with the predicted probabilities reflecting the confidence level for each category. Based on these probability outputs, the feature sequences are sorted and scanned in descending order, allowing low-confidence features to learn embeddings from high-confidence features. We also provide theoretical proof demonstrating that this probability-guided scanning method offers superior feature learning compared to random scanning. Notably, our method enables direct training on large-scale images without data splitting, eliminating the patch misalignment issue. 

In summary, the contributions of this paper are four folds:
\begin{itemize}
    \item [(1)] We propose a novel category prompt Mamba block in nuclei segmentation and classification network, which enables direct network training on large-scale images without requiring a data splitting process.
    \item [(2)] We design a patch-level category prompt method, which employs multi-class labels as supervision information to help the network learn class-related features.
    \item [(3)] We introduce a probability-guided sorting and scanning strategy to enhance the representation of rare categories.  We also provide theoretical proof of the sorting method.
    \item [(4)] We conduct extensive experiments on the four nuclei segmentation and classification datasets, achieving state-of-the-art (SOTA) performance while significantly improving training efficiency.
\end{itemize}

\section{Related Work}

\subsection{Nuclei Segmentation and Classification}

In recent years, deep learning has revolutionized nuclei analysis by eliminating the need for complex threshold selection \cite{win2017automated} and feature extraction processes \cite{xu2016automatic, lou2024structure}. Current research in nuclei segmentation focuses on addressing the challenge of boundary overlap. For instance, DCAN \cite{chen2016dcan} introduces a separate contour prediction branch to enhance segmentation accuracy. Methods like HoverNet \cite{graham2019hover}, and Dist \cite{naylor2018segmentation} utilize a distance regression branch to improve edge discrimination. Similarly, CDNet \cite{he2021cdnet} and SONNET \cite{doan2022sonnet} discretize the original distance map to refine segmentation outcomes. Additional methods continue to advance the field of nuclear segmentation \cite{ahmad2023dan, he2023transnuseg}. In the domain of nuclei classification, several models have been proposed based on graph neural networks (GNNs). For example, MPNet \cite{hassan2022nucleus} introduces a message-passing network to aggregate global information. While EAGNN \cite{hasegawa2023edge} incorporates edge labels into the GNN architecture. SENC \cite{lou2024structure} enhances nuclei representation through nuclear structure learning, and CGT \cite{lou2024cell} combines GNNs with Transformers to establish graph edge relationships. Despite these advancements, imbalanced nuclear category distribution remains a significant hurdle. This paper introduces a novel category prompt network designed to enhance classification accuracy under imbalanced scenarios.

\begin{figure*}[t!]
	\centering
	\includegraphics[width=7.0in]{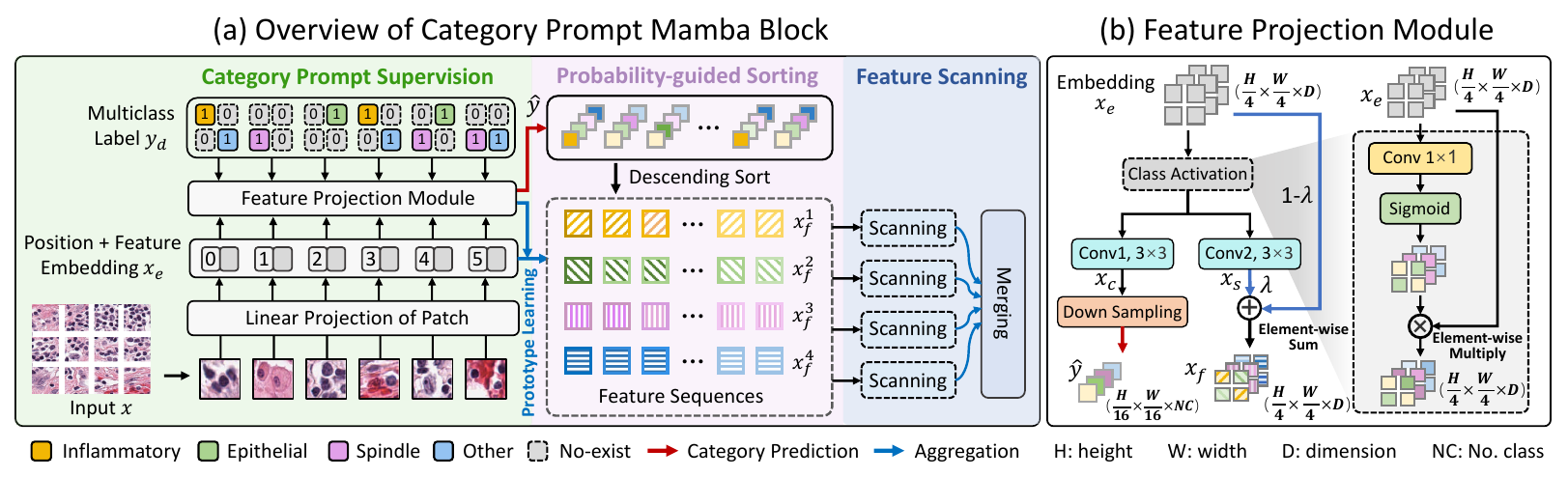}
	\caption{\textbf{The framework of our proposed CP-Mamba.} The method employs the category supervision information to learn each type nuclear prototype. Meantime, the probability predictions provide the guides for feature sequence ordering. }
	\label{fig: flow chart}
        \vspace{-2mm}
\end{figure*}

\subsection{Mamba in Medical Image Analysis}

The state space model (SSM) is well-regarded for its ability to capture long sequence dependencies with linear time complexity. Among the SSM models, Mamba \cite{gu2023mamba} gains significant attention due to its state-space selection mechanism, which allows for dynamic adjustment of model parameters based on the input. This adaptability leads to the widespread adoption of Mamba-based models in various domains, including medical image analysis. Several models have successfully integrated Mamba as a core component. For instance, U-Mamba \cite{ma2024u} and Mamba-UNet \cite{wang2024mamba} utilize Mamba as a feature extractor.
Additionally, models like Vim \cite{zhu2024vision}, and VMamba \cite{liu2024vmamba} introduce bidirectional feature scanning and four-direction cross-scanning techniques, respectively, to enhance the feature extraction capabilities of the encoder. These approaches demonstrate effectiveness in various tasks, including semantic and instance segmentation. Further advancements in sequence scanning mechanisms are made with methods such as Zigzag Scan \cite{hu2024zigma}, Omnidirectional Selective Scan \cite{shi2024vmambair}, and Hierarchical Scan \cite{zhang2025motion}. These techniques aim to optimize the scanning process for better feature representation. In the field of multiple instance learning, Mamba-based methods \cite{fang2024mammil, yang2024mambamil} are proposed to leverage Mamba's sequence modeling capabilities, enhancing the interaction between instances. 
In this paper, we introduce a nuclei segmentation and classification network, which allows for direct training on large-scale images, eliminating the need for patch-based image splitting processing.

\section{Methods}
\subsection{Overview}

In this work, we propose a category prompt Mamba (CP-Mamba) as encoder block for feature extraction. Leveraging Mamba's low memory usage and long-sequence modeling capabilities, our approach enables direct training on large-scale images, eliminating the need for image splitting. To tackle the class imbalance challenge, a category phenotype learning module and a probability-guided sorting strategy are designed employing category prompt information to enhance the network's ability to represent class-specific features. The overall framework is illustrated in Fig. \ref{fig: flow chart}.

\begin{algorithm}[t!]
\caption{Multi-class Labels Generation}
\label{alg:downsampling}
\begin{algorithmic}[1]
\STATE \textbf{Input:} Nuclei Ground-truth Classification Mask: $y$;
\STATE \textbf{Output:} Down-sampled Multi-class Label: $y_d$;
\STATE Calculating the Dimension of $y$: $h, w \leftarrow Shape(y)$;
\STATE \# Down-sampling scale 16 
\STATE Initializing the Multi-class Label: $y_d\in\{0\}^{\frac{h}{16}\times \frac{w}{16} \times NC}$ ;
\FOR{$i \in\{1, \cdots, NC\}$}
\FOR{$h', w' \in \{0, 16, 32, 48, \cdots\}$}
\STATE if $i \in y_{[h':h'+16, w':w'+16]}$:
\STATE \quad \quad $y_{d[h'/16, w'/16, i]} = 1$
\ENDFOR
\ENDFOR
\STATE \textbf{return} $y_d$ \quad \# Size($y_d$)=($\frac{h}{16}\times \frac{w}{16}\times NC$)
\end{algorithmic}
\end{algorithm}

\subsection{Category Prompt Supervision}

Traditional pixel-by-pixel classification methods often struggle with class-imbalanced instance segmentation tasks. These methods treat each pixel equally, overlooking the class distribution. Previous studies \cite{yue2024surgicalsam} suggest that category information, as a weakly supervised signal, can help the network learn positive activation features, thereby improving segmentation and classification performance. To enhance the network's perception of nuclei class distribution and improve segmentation performance, we design a category prompt supervision method based on patch-level category information, as illustrated in Fig. \ref{fig: flow chart} (a).

Given an input image $x \in \mathcal{R}^{H \times W \times C}$, we first reshape it into a 2D patch sequence $x_p \in \mathcal{R}^{N \times (P^2 \cdot C)}$, where $(H, W)$ represents the image resolution, $C$ is the number of channels, $N$ is the number of patches, and $P$ is the patch size. For this process, we set the patch size to $4\times4$. The patch sequences are then fed into the network, where they are concatenated with positional embeddings to generate feature embeddings $x_e=[f(x_p), x_{pos}]$, where $f$ denotes the linear projection layers.
Next, we designed a category prompt supervision method. This method uses the multi-class labels to supervise feature learning, with a key component being the feature projection module shown in Fig. \ref{fig: flow chart} (b). In this module, $x_e$ is input into the feature projection module to obtain a new class-related feature representation $x_c$:
\begin{equation}
    x_c = Conv_1(CA(x_e)), 
\end{equation}
where $Conv_1$ denotes a $3\times 3$ convolution layer, and $CA$ represents class activation layers consisting of a $1\times 1$ convolution layer followed by a sigmoid activation function. The $1 \times 1$ convolution layer and activation function allow the network to activate features corresponding to specific classes selectively. By multiplying these activated features with the original features, the network can effectively extract and enhance the feature representation for each class.

Predicting category labels directly on a $4 \times 4$ patch presents significant challenges. Such a small patch size is insufficient to encompass an entire nucleus, making it challenging to capture the semantic features necessary for accurate classification. Additionally, using smaller patches increases computational complexity due to the more significant number of patches involved in the loss calculation. To address these issues, we apply a down-sampling operation to the output $x_c$, as illustrated on the left side of Fig. \ref{fig: flow chart} (b):
\begin{equation}
    \hat{y} = DS(x_c),
\end{equation}
where $DS$ denotes the down-sampling operation with a scale factor of 4. This ensures that each pixel in the feature map $\hat{y}$ corresponds to a $16\times16$ region in the original input $x$, comparable to the size of a nucleus. After obtaining $\hat{y}$, we compute the multi-label classification loss using the category prompt labels $y_d$, derived from the ground-truth nuclei classification label $y$. The generation process of $y_d$ is detailed in Algorithm 1. The loss function is defined as:
\begin{equation}
    L_{p} = MCE(y_{d},\hat{y}),
\end{equation}
where $MCE$ represents the multi-class cross-entropy loss.

\subsection{Category Phenotype Learning}

Following a similar process to the category prompt supervision, the feature $x_e$ is input into the CA module followed by a $3 \times 3$ convolution layer, obtaining a new category phenotype representation $x_s$:
\begin{equation}
    x_s = Conv_2(CA(x_e)).
\end{equation}
In this procedure, since $x_s$ shares the same class activation module as the category prompt supervision, $x_s$ primarily contains class-related semantic features. To better capture feature embeddings relevant to segmentation and classification tasks, we design a feature fusion method as shown in the right branch of Fig. \ref{fig: flow chart} (b). In the design, $x_s$ is fused with feature $x_e$ to consider the class-independent semantic features from the original input. The fused feature will be used for sequence sorting. The fusion method is shown as follows:
\begin{equation}
    x_f = \lambda \cdot x_s + (1-\lambda) \cdot x_e,
\end{equation}
where $\lambda$ is a weighting parameter, which combines the original segmentation-related information with the class-related semantic information, enhancing the feature representation $x_f$. In this paper, we set $\lambda$ to 0.2, and more parameter ablation experiments are provided in \textbf{supplementary materials}.

\begin{figure}[t!]
    \centering
    \includegraphics[width=3.2in]{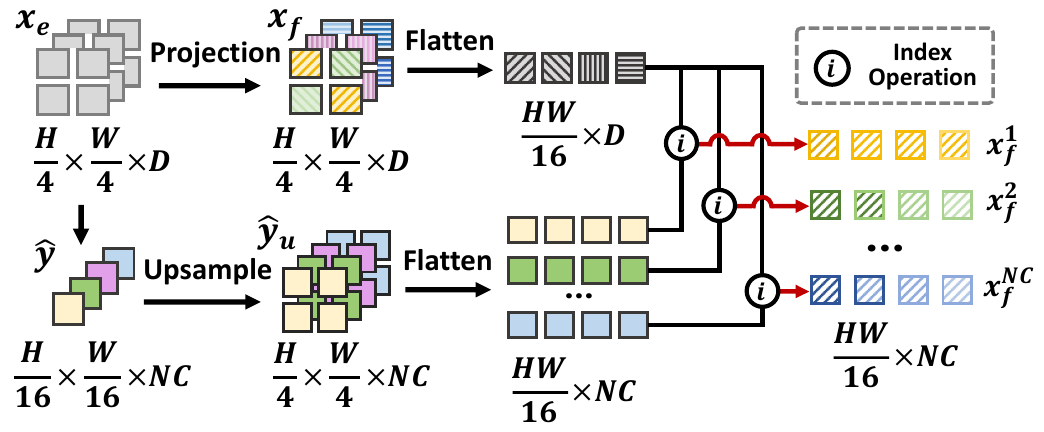}
    \caption{\textbf{The category sorting method guided by probability prediction.} ``i'' represents the sorting by index.}
    \label{fig: sort}
    \vspace{-5mm}
\end{figure}

\subsection{Probability-guided Sequence Sorting}
Previous Mamba network scanning methods are direction-based \cite{wang2024mamba, liu2024vmamba}, which may not effectively handle class-imbalanced samples. When only a few patches contain a specific category, the distance between these patches can hinder the network's ability to extract features for that category. To address this, we propose a probability-guided sequence sorting method that encourages low-confidence features to learn from high-confidence ones. Hence, this sorting mechanism allows preceding patches to provide category-prior knowledge for subsequent patches, reducing classification uncertainty.

Guided by the probability prediction $\hat{y}$, we implement the probability-guided sorting method shown in Fig. \ref{fig: sort}. Since $\hat{y}$ is the output of the down-sampling operation and does not align with $x_f$ in the feature dimension, we first up-sample $\hat{y}$ to generate $\hat{y}_u$. Next, we reshape $x_f$ and $\hat{y}_u$ to dimensions $\frac{HW}{16} \times D$ and $\frac{HW}{16} \times NC$, respectively, where $D$ is the feature dimension and $NC$ is the number of categories. 
We then sort flattened sequences from highest to lowest based on the $i^{th}$ class probability ordering and generate sorted feature sequence $\{x_f^{i}, i = 1,\cdots, NC\}$ for each category. In the end, these sequences are utilized for feature scanning, which is detailed in the following subsection. To prove the effectiveness of the sorting method, we give the following theorem.

\noindent \textbf{Theorem} \textit{Given a random sorting feature sequence $X_{seq}=\{X_1, X_2, \cdots, X_n\}$, we use the joint entropy $H(X_{seq})$ to represent the classification uncertainty when $X_{seq}$ used as classification features. When the sequence is sorted according to task-related confidence level from highest to lowest, then the joint entropy of the sequence will decrease, i.e.}
$H(X_1^{'}, X_2^{'}, \cdots, X_n^{'}) \leq H(X_1, X_2, \cdots, X_n)$. 

Where $\{X_1^{'}, X_2^{'}, \cdots, X_n^{'}\}$ represents a sorted sequence according to classification confidence level from highest to lowest. This theorem implies that probability-guided sorting can decrease the uncertainty of classification tasks and the proof of the theorem is provided in the \textbf{supplementary materials}. In addition, we conducted ablation experiments to analyze the effectiveness of the sorting method.

\begin{table*}[t!]
\centering
\renewcommand{\arraystretch}{1.1}
	\resizebox{\linewidth}{!}{
    \begin{tabular}{clccccc|clccccc|c}
        \Xhline{0.8pt}
        Datasets & Methods & DICE & AJI & DQ & SQ & PQ & Datasets & Methods & DICE & AJI & DQ & SQ & PQ & Publications \\
        \Xhline{0.8pt}
        & Mask-RCNN$^\dag$ & 74.59 & 60.66 & 75.04 & 73.96 & 56.15 & & Mask-RCNN$^\dag$ & 74.96 & 51.88 & \textbf{66.11} & 75.73 & 49.81 & ICCV'2017\\
        & HoverNet$^\ddag$ & \underline{80.76} & \underline{64.21} & \underline{78.66} & \textbf{76.54} & \underline{61.49} & & HoverNet$^\dag$ & \underline{81.97} & \underline{53.64} & 64.14 & \underline{76.29} & \underline{50.38} & MIA'2019\\
        GLySAC & Triple-UNet$^\ddag$ & 75.27 & 62.03 & 76.15 & 74.10 & 57.87 & CoNSeP & Triple-UNet$^\ddag$ & 80.39 & 39.25 & 49.81 & 74.66 & 37.25 & MIA'2020 \\
        & DoNet$^\dag$ & 75.22 & 61.89 & 74.34 & 74.21 & 56.99 &  & DoNet$^\dag$ & 78.23 & 46.76 & 57.27 & 72.89 & 45.64 & CVPR'2023 \\
        & Vim$^\ddag$ & 79.82 & 63.11 & 77.29 & 75.02 & 59.49 &  & Vim$^\ddag$ & 80.64 & 49.21 & 62.88 & 75.94 & 49.80 & ICML'2024 \\
        & \textbf{Ours}$^\ddag$	& \textbf{81.43} & \textbf{64.87} & \textbf{79.02} & \underline{75.34} & \textbf{61.77} & & \textbf{Ours}$^\ddag$ & \textbf{82.23} & \textbf{53.90} & \underline{65.81} & \textbf{77.67} & \textbf{51.46} & - \\
        \Xhline{0.8pt}
        Datasets & Methods & DICE & AJI & DQ & SQ & PQ & Datasets & Methods & DICE & AJI & DQ & SQ & PQ & Publications \\
        \Xhline{0.8pt}
        & Mask-RCNN$^\dag$ & 73.23 & 60.80 & 74.18 & 78.27 & 59.44 & & Mask-RCNN$^\dag$ & 75.06 & 61.24 & 72.34 & 78.69 & 60.19 & ICCV'2017 \\
        & HoverNet$^\ddag$ & 74.41 & 61.27 & \textbf{75.88} & \underline{79.61} & \underline{60.47} & & HoverNet$^\ddag$ & \underline{80.36} & 65.64 & \underline{75.82} & 80.19 & \underline{61.22} & MIA'2019 \\
        MoNuSAC & Triple-UNet$^\dag$ & 50.84 & 43.18 & 62.15 & 67.67 & 39.65 & PanNuke & Triple-UNet$^\dag$ & 74.24 & 58.83 & 66.71 & 73.58 & 54.06 & MIA'2020 \\
        & DoNet$^\dag$ & 70.03 & 59.45 &  70.90 & 75.30 & 58.49 & & DoNet$^\dag$ & 78.24 & \underline{66.86} & 74.90 & 79.29 & 59.91 & CVPR'2023\\
        & Vim$^\ddag$ & \underline{74.48} & \underline{62.11} & 74.97 & 76.77 & 58.23 & & Vim $^\ddag$ & 79.69 & 64.89 & 75.59 & \underline{81.22} & 60.89 & ICML'2024\\
        & \textbf{Ours}$^\dag$	& \textbf{75.41} & \textbf{62.76} & \underline{75.40} & \textbf{80.08} & \textbf{60.68} & &  \textbf{Ours}$^\dag$ & \textbf{81.87} & \textbf{67.90} & \textbf{76.79} & \textbf{81.90} & \textbf{61.49} & -\\
    \Xhline{0.8pt}
    \end{tabular}}
    \caption{The nuclei segmentation comparison with the state-of-the-art methods on the GlySAC, ConSeP, MoNuSAC and PanNuke datasets. $\dag$ represents p-value of AJI $<$ 0.001 and $\ddag$ represents p-value of AJI $<$ 0.05.}
    \label{tab:seg_comp}
\end{table*}
\subsection{Network Training }

Our training network follows an encoder-decoder architecture. The encoder consists of four CP-Mamba blocks, each consisting of a category phenotype learning process described previously and a feature aggregation process. In the feature aggregation process, we adopt summary operation as VMamba \cite{liu2024vmamba}, which can be detailed as follows:
\begin{equation}
x_{enc} = \sum_{i=1}^{NC} S(x_f^i),
\end{equation}
where $S$ represents the feature scanning operation as Mamba \cite{gu2023mamba}. During the encoding process, category prompt supervision and probability-guided sorting facilitate patch interaction and allow for the independent extraction of class-related features. Hence, they strengthen the representation of rare classes. The detailed training architecture is provided in the \textbf{supplementary materials}.

The decoder comprises two parallel U-Net decode branches. The first branch is designed to learn the foreground, background, and contour semantic features, while the second branch focuses on the classification task. Instance segmentation is then derived from the post-processing operation. The overall training loss is defined as:
\begin{equation}
L = L_{p} + \alpha L_{sem} + \beta L_{cls},
\end{equation}
where $L_{sem}$ represents the loss of semantic branch, $L_{cls}$ represents the classification loss, and $L_{p}$ represents the loss of the category prompt supervision. The parameters $\alpha$ and $\beta$ balance the loss weight. We set $\alpha$ and $\beta$ to 1 in the paper. In $L_{sem}$ and $L_{cls}$, we simultaneously employ cross-entropy and dice losses to optimize the objective.

\section{Experiments}
\subsection{Datasets}

We evaluate our proposed model on four pathological datasets, including GLySAC \cite{doan2022sonnet}, CoNSeP \cite{graham2019hover}, MoNuSAC \cite{verma2021monusac2020}, and PanNuke \cite{gamper2019pannuke} datasets.
The \textbf{GLySAC} includes 59 H\&E stained images of size 1000$\times$1000 pixels from 8 gastric adenocarcinoma WSIs digitized at 40$\times$magnification. The dataset includes 30975 annotated nuclei grouped into three categories: lymphocytes, epithelial, and miscellaneous.
The \textbf{ConSeP} involves 41 H\&E stained images of size 1000$\times$100 pixels. The dataset includes 24319 annotated nuclei grouped into four types: miscellaneous, inflammatory, epithelial, and spindle.
The \textbf{MoNuSAC} comprises 209 annotated images, ranging from 81$\times$113 pixels to 1422$\times$2162 pixels. The dataset comprises 31411 annotated nuclei of four categories: epithelial nuclei, lymphocytes, macrophages, and neutrophils.
The \textbf{PanNuke} includes 7899 images of size 256$\times$256 pixels obtained from 19 organs. The dataset contains 189744 annotated nuclei from five categories, including neoplastic, non-neoplastic epithelial, inflammatory, connective, and dead nuclei.

\subsection{Implementation Details and Evaluation Metrics}

Our all experiments are run with PyTorch on two Nvidia RTX 4090 GPUs. We use SGD as an optimizer, and the learning rate, momentum, and weight decay are set at 0.01, 0.9, and 0.0005. Besides, we train the network for 6000 iterations. We evaluate the segmentation performance over metrics of DICE, AJI \cite{kumar2017dataset}, DQ \cite{kirillov2019panoptic}, SQ, and PQ and evaluate the classification performance with an F1 score. Throughout all the tables in this paper, we bold the \textbf{best} and underline the \underline{second best}.

\begin{figure}[t]
	\centering
	\includegraphics[width=3.0in]{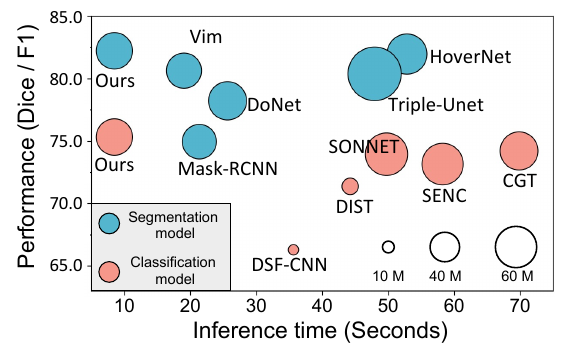}
	\caption{\textbf{The model complexity analysis.} The ``green" bubbles represent the segmentation model, and the ``red" bubbles represent the classification model. The size of the bubble represents the size of the model parameters.}
	\label{fig:complex}
        \vspace{-3mm}
\end{figure}

\begin{figure*}[t!]
	\centering
	\includegraphics[width=5.5in]{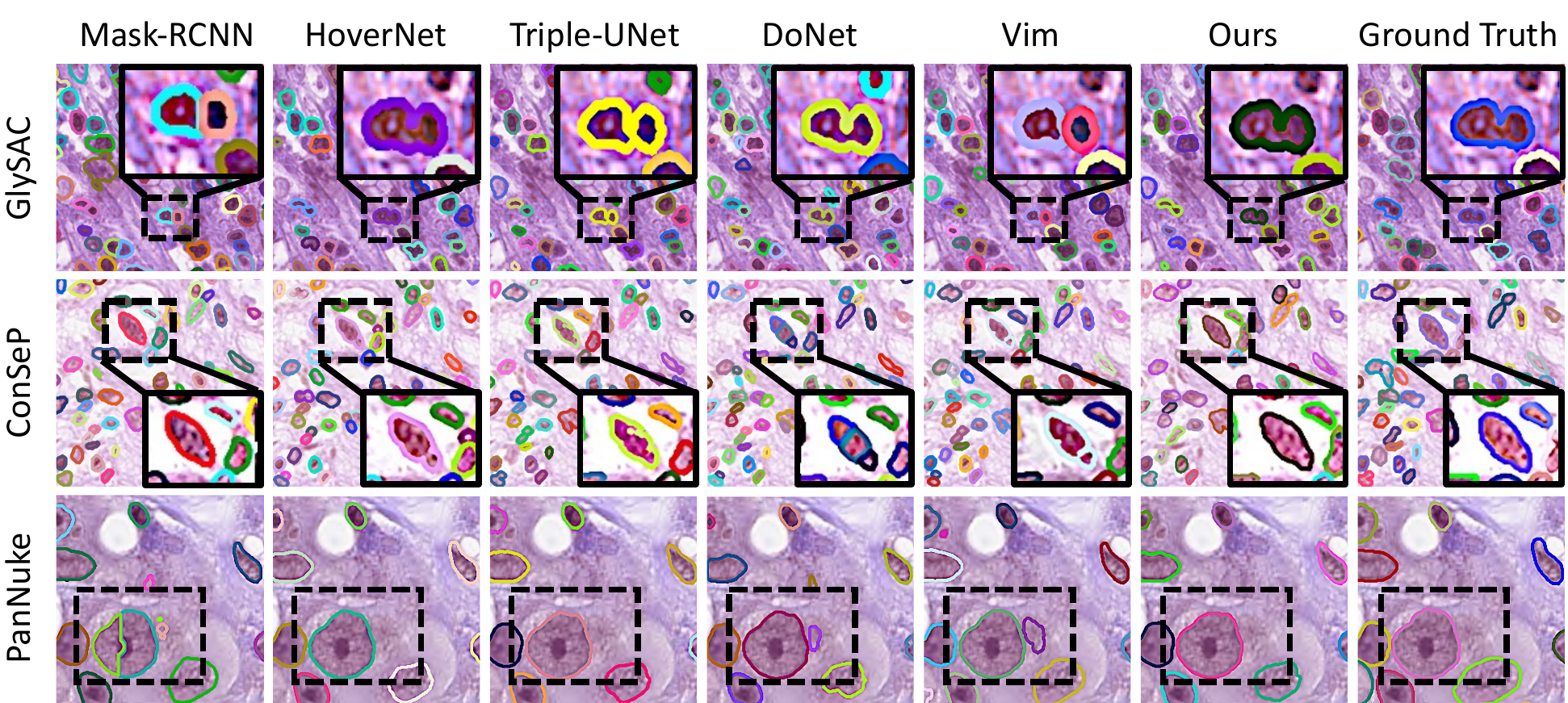}
	\caption{\textbf{The visualization comparison on the segmentation task.} The black boxes highlight the segmentation differences.}
	\label{fig:visual_comp3}
\end{figure*}

\begin{table*}[t!]
\centering
\renewcommand{\arraystretch}{1.1}
\resizebox{\linewidth}{!}{
    \begin{tabular}{clccccc|clcccccc|c}
        \cline{1-16}
        Datasets & Methods & $F_d$ & $F^E$ & $F^L$ & $F^M$ & - & Datasets & Methods & $F_d$ & $F^E$ & $F^I$ & $F^M$ & $F^S$ & - & Publications \\
        \cline{1-16}
        & DIST$^\dag$ & 80.11 & 50.81 & 50.67 & 16.92 & - & & DIST$^\dag$ & 71.39 & 61.65 & 60.89 & 17.18 & 52.67 & - & TMI'2018 \\
        & DSF-CNN$^\dag$ & 82.79 & 52.64 & 49.76 & 25.27 & - &  & DSF-CNN$^\dag$ & 66.31 & 56.79 & 54.17 & 11.97 & 9.32 & - & TMI'2020\\
        GLySAC & SONNET$^\ddag$ & 83.29 & 52.57 & 51.76 & 32.68 & - & ConSeP & SONNET$^\ddag$ & 73.96 & 64.03 & \underline{61.96} & 36.78 & \underline{55.98} & - & JBHI'2022 \\
        & CGT$^\ddag$ & \underline{85.76} & \underline{55.23} & \underline{52.61} & 34.94 & - & & CGT$^\ddag$ & \underline{74.22} & \underline{64.68} & 56.23 & \underline{39.66} & 54.17 & - & AAAI'2024 \\
        & SENC$^\ddag$ &  84.59 & 55.08 & 51.73 & \underline{35.10} & - &  & SENC$^\ddag$ & 73.17 & 60.60 & 58.16 & 39.47 & 55.73 & - & MIA'2024 \\
        & \textbf{Ours}$^\ddag$	& \textbf{86.94} & \textbf{56.15} & \textbf{53.49} & \textbf{35.92} & -	& &  \textbf{Ours}$^\ddag$ & \textbf{75.33} & \textbf{65.78} & \textbf{63.28} & \textbf{40.96} & \textbf{57.17} & - & -\\
        \cline{1-16}
        Datasets & Methods & $F_d$ & $F^E$ & $F^L$ & $F^{Ma}$ & $F^N$ & Datasets & Methods & $F_d$ & $F^C$ & $F^{D}$ & $F^I$ & $F^{Ne}$ & $F^{No}$ & Publications\\
        \cline{1-16}
        & DIST$^\dag$ & 60.48 & 60.82 & 72.66 & 14.51 & 29.18 & & DIST$^\dag$ & 71.80 & 41.90 & 2.15 & 42.63 & 50.00 & 34.17 & TMI'2018 \\
        & DSF-CNN$^\dag$ & 82.11 & 79.18 & 75.41 & 40.19 & 50.80 & & DSF-CNN$^\dag$ & 78.28 & 44.94 & 6.89 & 47.20 & 59.34 & 50.09 & TMI'2020 \\
        MoNuSAC & SONNET$^\ddag$ & \textbf{83.63} & \textbf{83.48} & 78.57 & \underline{45.73} & \underline{57.19} & PanNuke & SONNET$^\ddag$ & \underline{80.02} & 45.34 & \underline{29.85} & \underline{53.17} & \underline{61.38} & 55.81 & JBHI'2022 \\
        & CGT$^\ddag$ & 81.59 & 80.44 & 79.67 & 41.64 & 52.08 & & CGT$^\ddag$ & 78.68 & 47.27 & 28.96 & 52.67 & 59.80 & 62.13 & AAAI'2024 \\
        & SENC$^\ddag$ & 75.82 & 78.19 & \underline{81.66} & 39.23 & 49.22 & & SENC$^\ddag$ & 76.78 & \underline{48.23} & 21.64  & 49.88  & 60.03 & \textbf{65.30} & MIA'2024 \\
        & \textbf{Ours}$^\ddag$ & \underline{82.14} & \underline{82.90} &  \textbf{82.89} & \textbf{46.17} & \textbf{58.23} & & \textbf{Ours}$^\ddag$ & \textbf{81.70} & \textbf{50.04} & \textbf{31.30} & \textbf{53.93} & \textbf{62.16} & \underline{64.08} & -\\
    \Xhline{1.0pt}
    \end{tabular}}
    \caption{The nuclei classification comparison with the state-of-the-art methods on the GLySAC, ConSeP, MoNuSAC and PanNuke datasets. $\dag$ represents p-value of $F_d<$0.001 and $\ddag$ represents p-value of $F_d <$ 0.05. $F^C$, $F^D$, $F^E$, $F^I$, $F^L$, $F^M$, $F^{Ma}$, $F^N$, $F^{Ne}$, $F^{No}$, and $F^S$ denote the F1 score for nuclear types of connective, dead, epithelial, inflammatory, lymphocyte, miscellaneous, macrophages, neutrophils, neoplastic, non-neo plastic, and spindle, respectively.}
    \label{tab: compare}
\end{table*}

\subsection{Computational Complexity}

To validate the superiority of our method in inference time, we first perform a complexity analysis as shown in Fig. \ref{fig:complex}, which shows the inference time, model parameters, and performance comparisons between our proposed method and other segmentation and classification models. We fix the input image size in this experiment as $1000 \times 1000$. In detail, segmentation models include Mask-RCNN \cite{he2017mask}, HoverNet \cite{graham2019hover}, Triple-UNet \cite{zhao2020triple}, DoNet \cite{jiang2023donet} and Vim \cite{zhu2024vision}. The classification models include DIST \cite{naylor2018segmentation}, DSF-CNN \cite{graham2020dense}, SONNET \cite{doan2022sonnet}, CGT \cite{lou2024cell}, SENC \cite{lou2024structure}. The figure shows that our proposed method is significantly superior in classification and segmentation tasks.
On the one hand, our model has less inference time because our method directly inferences on the large images. In contrast, other methods consist of an image splitting step, which enhances the inference time by linear time. On the other hand, our method can achieve better performance under less parameters, and the detailed performance comparison will be shown in the following subsections.

\subsection{Comparison with the State-of-the-art Methods}
\subsubsection{Segmentation Performance}

We validate the effectiveness of our method with five state-of-the-art nuclei segmentation models on four datasets. These comparison methods include Mask-RCNN \cite{he2017mask}, HoverNet \cite{graham2019hover}, Triple-UNet \cite{zhao2020triple}, DoNet \cite{jiang2023donet} and Vim \cite{zhu2024vision} and comparison results are shown in Table \ref{tab:seg_comp}. 
The results show that our method performs best in Dice and PQ metrics. In addition, compared with the second-best method, HoverNet, our method still significantly improved with an AJI increase of 0.6, 0.3, 0.5, and 0.2 on four datasets, respectively. Combined with the complexity analysis experiment of the previous subsection, the comparison results show that our method does not damage the segmentation performance and dramatically reduces the inference time when training on large-scale images.

\begin{figure*}[t]
	\centering
	\includegraphics[width=6.7in]{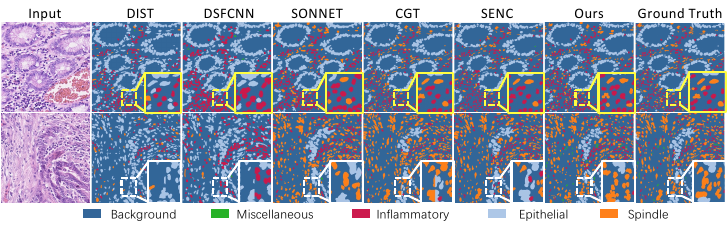}
	\caption{\textbf{The visualization comparison on the classification task.} The yellow boxes and white boxes highlight the misclassified and not-aligned nuclei respectively.}
	\label{fig:comp1}
        \vspace{-2mm}
\end{figure*}

Furthermore, the visualization comparisons of nuclei segmentation on GLySAC, ConSeP and PanNuke datasets are shown in Fig. \ref{fig:visual_comp3}. From the results we can see our prediction outputs are closer to the ground-truth labels. In detail, Mask-RCNN and Vim mistakenly divide a single nucleus into multiple ones. Meanwhile, DoNet and Vim incorrectly identify tissue regions as nuclei.

\subsubsection{Classification Performance}

We compare the classification performance of our method with five SOTA classification networks, namely DIST \cite{naylor2018segmentation}, DSF-CNN \cite{graham2020dense}, SONNET \cite{doan2022sonnet}, CGT \cite{lou2024cell}, and SENC \cite{lou2024structure}. The comparison performances are shown in Table \ref{tab: compare}. In this table, $ F_d$ represents the F1 score of the detection, and the others represent the F1 score of classification of each type. Overall, our method's classification performance is better, especially in the face of an unbalanced distribution of categories. In the ConSeP dataset, the $F^{I}$ scores of DIST and DSF-CNN on miscellaneous nuclei are only 17.18\% and 11.97\%, and our method has increased by more than 20\%. Similar comparison results can be drawn from the MoNuSAC dataset. Such as, when facing rare miscellaneous and neutrophil nuclei, our method still performs best and beyond Dist around 30\%.

Furthermore, we provide classification visualizations on the $1000 \times 1000$ images from the ConSeP in Fig. \ref{fig:comp1}. In the experiment, we use ground-truth masks as segmentation results for CGT and SENC because they only execute classification tasks and do not have segmentation predictions. From the figure, our method performs better when facing spindle nuclei in yellow boxes. However, DIST and DSFCNN usually misclassified spindles into epithelial or inflammatory. In the meantime, according to the enlarged white areas, we can see methods employing image clipping cannot be aligned between the nucleus, such as DSFCNN and CGT. In contrast, our method has better nuclear integrity.

\begin{table}[t!]
    \centering
    \renewcommand{\arraystretch}{1.3}{
    \resizebox{\linewidth}{!}{
    \begin{tabular}{c|ccc|cccc}
    \cline{1-8}
    \multirow{2}{*}{Scanning Direction} & \multicolumn{3}{c}{GlySAC} & \multicolumn{3}{c}{ConSeP} \\
    \cline{2-8}
     & $F^E$ & $F^L$ & $F^M$ & $F^E$ & $F^I$ & $F^M$ & $F^S$\\
    \cline{1-8}
    Bidirectional & 53.65 & 53.12 & 29.27 & 64.49 & 49.14 & 25.33 & 56.86\\
    Cross-Scanning & 54.34 & 53.16 & 31.08 & 65.19 & 62.79 & 34.13 & 55.73\\
    Probability Sorting & \textbf{56.15} & \textbf{53.49} & \textbf{35.92} & \textbf{65.78} & \textbf{63.28} & \textbf{40.96} & \textbf{57.17} \\
    \cline{1-8}
    \end{tabular}}}
    \caption{The effect of scanning direction for nuclei classification on GlySAC and ConSeP datasets.}
    \label{tab:ab1}
\end{table}

\subsection{Ablation Study}

To validate the effect of our probability sorting strategy on resolving class-imbalance problems, we compared the different scanning strategies on GLySAC and ConSeP datasets, including bidirectional scanning and cross-scanning. As seen from Table \ref{tab:ab1}, our probability-guided sorting method has a significant improvement, especially in the miscellaneous, the bidirectional scanning method is 6\% and 15\% lower than our method, and the cross-scanning method is 4\% and 6\% lower than our method. The above results show that probability sorting vastly improves the classification performance of rare categories. Next, we conduct ablation analysis for the model design, including the probability sorting strategy and phenotype learning, and the results are shown in Table \ref{tab:ab2}. Overall, when either of the two designs is removed, the model's performance is significantly reduced, indicating that the design of both modules is effective. In detail, when phenotype learning is not used, the classification performance decreases by more than 6\% and 3\% on the two datasets, respectively, indicating that the aggregated class-related features from the category prompt supervision are practical for model classification.

\begin{table}[t!]
    \centering
    
    \renewcommand{\arraystretch}{1.15}{
    \resizebox{\linewidth}{!}{
    \begin{tabular}{cc|ccc|ccc}
    \cline{1-8}
    \multirow{2}{*}{Sorting} & \multirow{2}{*}{Phenotype} & \multicolumn{3}{c}{GlySAC} & \multicolumn{3}{c}{ConSeP} \\
    \cline{3-8}
     & & Dice & AJI & $F_d$ & Dice & AJI & $F_d$ \\
    \cline{1-8}
     & & 79.27 & 61.14 & 79.42 & 78.34 & 51.07 & 71.12\\
    \ding{51} & & 80.09 & 62.82 & 80.20 & 79.98 & 52.21 & 72.22 \\
    & \ding{51} &  79.79 & 63.34 & 84.75 & 80.09 & 52.23 & 73.83 \\
    \ding{51} & \ding{51} & \textbf{81.43} & \textbf{64.87} & \textbf{86.94} & \textbf{82.23} & \textbf{53.90} & \textbf{75.33} \\
    \cline{1-8}
    \end{tabular}}}
    \caption{Ablation experiments of module designing on GlySAC and ConSeP datasets.}
    \label{tab:ab2}
    \vspace{-2mm}
\end{table}

\section{Conclusions}

In this paper, we propose a probability-guided sorting method based on Mamba to solve the task of nuclei segmentation and classification in the case of class imbalance. This method uses category prompts to generate confidence probability for each category, and the prediction results are used as reference guide sequence sorting. Then, through independent feature scanning and aggregation of each category sequence, the method boosts the feature representation of rare categories to improve classification accuracy. In addition, our method can be trained directly on large-scale images without an image splitting process, which not only solves the problem of the nuclei not being aligned with the traditional method but also dramatically reduces the training and inference time of the model. We perform extensive comparative experiments on four datasets, and the experiment results also demonstrate the validity of our proposed method.

\section{Acknowledgments}

This work was supported in part by the National Natural Science Foundation of China under 62031023 \& 62331011, in part by the Shenzhen Science and Technology Project under GXWD20220818170353009, and in part by the Fundamental Research Funds for the Central Universities under No.HIT.OCEF.2023050.

\bibliography{aaai25}

\end{document}